\documentclass[lettersize,journal]{IEEEtran}
\usepackage{amsmath,amsfonts}
\usepackage{algorithmic}
\usepackage{algorithm}
\usepackage{array}
\usepackage[caption=false,font=normalsize,labelfont=sf,textfont=sf]{subfig}
\usepackage{textcomp}
\usepackage{stfloats}
\usepackage{url}
\usepackage{verbatim}
\usepackage{graphicx}
\usepackage{cite}
\usepackage{multirow}
\usepackage{multicol}
\usepackage{amsthm}
\usepackage{setspace}
\usepackage{bbding}
\begin{document}

\title{LightGCNet: A Lightweight Geometric Constructive Neural Network for Data-Driven Soft sensors}

\author{Jing Nan,~\IEEEmembership{Student Member,~IEEE}, Yan Qin,~\IEEEmembership{Member,~IEEE}, Wei Dai,~\IEEEmembership{Senior Member,~IEEE},  and Chau Yuen,~\IEEEmembership{Fellow,~IEEE}
\thanks{This work was supported in part by the National Key Research and Development Program of China under Grant 2022YFB3304700, the National Natural Science Foundation of China under Grant 62373361, and in part by the State Scholarship Fund, China Scholarship Council, under Grant 202306420127. (Corresponding author: Wei Dai)

Jing Nan is with the School of Information and Control Engineering, China University of Mining and Technology, Xuzhou, China 221116, and also with the Engineering Product Development Pillar, Singapore University of Technology and Design, Singapore 487372. e-mail: jingn@cumt.edu.cn.

Yan Qin is with the School of Automation, Chongqing University, and Engineering Product Development Pillar, The Singapore University of Technology and Design, 8 Somapah Road, 487372 Singapore. e-mail: yan.qin@cqu.edu.cn.

Wei Dai is with the School of Information and Control Engineering and Artificial Intelligence Research Institute, China University of Mining and Technology, Xuzhou, China 221116. e-mail: weidai@cumt.edu.cn.


Chau Yuen is with the School of Electrical and Electronics Engineering,
Nanyang Technological University, 50 Nanyang Ave, Singapore 639798.
email: chau.yuen@ntu.edu.sg.

}
\thanks{}}
\markboth{IEEE Transactions on Automation Science and Engineering}%
{Shell \MakeLowercase{\textit{et al.}}: LightGCNet: A Lightweight Geometric Constructive Neural Network for Data-Driven Soft sensors}


\maketitle

\begin{abstract}
Data-driven soft sensors provide a potentially cost-effective and more accurate modeling approach to measure difficult-to-measure indices in industrial processes compared to mechanistic approaches. Artificial intelligence (AI) techniques, such as deep learning, have become a popular soft sensors modeling approach in the area of machine learning and big data. However, soft sensors models based deep learning potentially lead to complex model structures and excessive training time. In addition, industrial processes often rely on distributed control systems (DCS) characterized by resource constraints. Herein, guided by spatial geometric, a lightweight geometric constructive neural network, namely LightGCNet, is proposed, which utilizes compact angle constraint to assign the hidden parameters from dynamic intervals. At the same time, a node pool strategy and spatial geometric relationships are used to visualize and optimize the process of assigning hidden parameters, enhancing interpretability. In addition, the universal approximation property of LightGCNet is proved by spatial geometric analysis. Two versions algorithmic implementations of LightGCNet are presented in this article. Simulation results concerning both benchmark datasets and the ore grinding process indicate remarkable merits of LightGCNet in terms of small network size, fast learning speed, and sound generalization.

\emph{Note to Practitioners}——Motivated by the conflict between the high resource requirements of the traditional data-driven soft sensors models and the generally inadequate resource of DCS, this article proposes a lightweight approach that offers significant advantages in terms of small network size and fast learning speed. In this work, we visualize the black-box modeling process of data-driven soft sensors models based on knowledge of spatial geometry. Preliminary experiments suggest that this approach is feasible.

\end{abstract}


\section{Introduction}
\IEEEPARstart{D}{ata-driven} soft sensors model is essentially an inferential model for key indices, constructed using artificial intelligence techniques and easy-to-measure indices\cite{ref1,ref2,ref3}. However, the performance of the soft sensors model is vulnerable to factors, such as time-varying operating conditions, data diversity, and conceptual drift\cite{ref4}. Therefore, the soft sensors model needs to be reconstructed to maintain the high inferential performance. It is well known that industrial processes typically rely on distributed control systems (DCS), which are often severely limited in terms of computational and storage capabilities\cite{ref5}. Model reconstruction consumes substantial computational resources and storage capacity, which may affect the normal operation of the DCS. Thus, the lightweight of the soft sensors model is particularly crucial for deployment on resource-constrained DCS devices.

Fortunately, single hidden layer feedforward network (SLFN) has been demonstrated to have great potential in developing low-cost and fast learning models\cite{ref6,ref7,ref8,ref9}. The SLFN is a lightweight flatted feedforward network with only one hidden layer. Its training consists of two subtasks: optimizing the network parameters (i.e., hidden parameters and output weights) and constructing the optimal hidden layer network architecture \cite{ref10}. Since the output weights have a linear nature, they can be efficiently obtained using the least square method in one step. Traditionally, the gradient descent-based approach and the trial-and-error approach are used to train the nonlinear parameters and determine the hidden layer network architecture, respectively\cite{ref11}. However, this approach may lead to poor model performance in terms of both generalizability and short modeling time, because nonlinear parameter training involves highly nonconvex landscapes and different trials are independent of each other. 

Considering these two subtasks, a combination method of randomized approaches (RA) and constructive algorithms (CA) was employed for training SLFN, i.e. constructive feedforward network with random weight (CFN-RW)\cite{ref12}. In CFN-RW, the hidden parameters are randomly generated by RA, and the network topology is dynamically determined by CA. However, the CFN-RW primarily emphasizes convergence accuracy, but ignores the numerical stability, which is an important indicator for the sensitivity of data perturbation\cite{ref13}. For CFN-RW, the numerical stability can be quantitatively characterized by the hidden layer output matrix which is determined by hidden parameters and input data. Therefore, it is apparent that only an elaborate selection of hidden parameters with respect to input data will lead to a CFN-RW with good performance.  Also, theoretical proofs show that the universal approximation property of the CFN-RW is conditional\cite{ref14}.

By analyzing the feasibility of CFN-RW for data modeling, a supervisory mechanism with data dependency is proposed to constrain the generation of hidden parameters\cite{ref15}. Due to the lack of flexible hyperparameters, the speed of network convergence is easily affected. Then, a constraint mechanism based on a non-negative sequence is proposed to accelerate the learning process\cite{ref16}. However, these constraints generate hidden parameters that may cause the saturation of the activation function, rendering it challenging to fully exploit its nonlinearity. Then, a hidden parameters generation method that fuses the input data and the activation function is proposed, which places the activation function in a randomly selected region of the input space and then selects the random parameters based on the local fluctuations of the objective function\cite{ref17}. Based on the above studies, a tighter inequality constraint (TIC) is proposed to improve the quality of hidden parameters\cite{ref18}. Although the remarkable ability of these constraints makes CFN-RW preferred as data-driven soft sensors model, it is still challenging to develop a lightweight model for industrial process. Here, we summarize that applying the CFN-RW soft sensors model in industrial process faces several challenges:

   1)
   \begin{minipage}[t]{230pt}
	Lack of theoretical foundation:	while above constraints significantly enhance the performance of CFN-RW, these studies have not thoroughly elucidated the fundamental principles behind the design constraints.
	\end{minipage}
   \setstretch{0.9}
   
   2)
   \begin{minipage}[t]{230pt}
	Lack of interpretability: the internal logic of these constraints is not easily comprehensible to humans, which is why they are commonly criticized as black-box constraints.
	\end{minipage}
	   \setstretch{0.9}
	   
To solve the above mentioned challenges, this article proposes a lightweight geometric constructive neural network, called LightGCNet. The main contribution of this article can be summarized as follows:

   1)
   \begin{minipage}[t]{230pt}
   Based on the spatial geometric relationship between hidden parameters, current residuals, and future residuals, a compact angular constraint is proposed for assigning hidden parameters.
   \end{minipage}
   \setstretch{0.9}
   
   2)
   \begin{minipage}[t]{230pt}
    Spatial geometric analysis is used to prove the universal approximation property of LightGCNet.
   \end{minipage}
   \setstretch{0.9}
   
   3)
   \begin{minipage}[t]{230pt}
    The process of assigning hidden parameters with compact angle constraint is visualized and optimized through node pool strategy and spatial geometric relationship.
   \end{minipage}
   \setstretch{0.9}
   
   4)
   \begin{minipage}[t]{230pt}
   Based on different evaluation approaches of output weights, two version algorithmic implementations, namely LightGCNet-I and LightGCNet-II, are proposed.
   \end{minipage}
   \setstretch{0.9}

The remainder of the article is organized as follows. Section II introduces the basic theory of CFN-RW and analyzes the CFN-RW. Section III proposes the LightGCNet and describes it in detail. Section IV reports the performance evaluation results. Finally, conclusions are drawn in section V.
\section{Problem Description}
\subsection{Constructive Feedforward Network with Random Weight}
CFN-RW can be regarded as a flatted network, where all the hidden parameters (the input weights and biases) are randomly assigned from a fixed interval and fixed during the training process. For a target function $f:{R^d} \to {R^m}$, assume that the CFN-RW with $L-1$ hidden nodes can be written as : ${f_{L - 1}} = {H_{L - 1}}{\beta ^ * }$, where ${H_{L - 1}} = \left[ {{g_1}\left( {w_1^T \cdot x + {b_1}} \right), \cdots ,{g_L}\left( {w_L^T \cdot x + {b_L}} \right)} \right]$, $T$ denotes matrix transpose, $x$ is the input sample, ${\omega _j}$ and ${b_j}$ are the input weights and biases of the $j$-th hidden node, respectively. $j = 1, \cdots ,L-1$, ${g_j}$ denotes the nonlinear activation function of the $j$-th hidden node. The output weights ${\beta ^ * } = {\left[ {{\beta _1},{\beta _2},...,{\beta _{L-1}}} \right]^T}$, where ${\beta _j}$ is evaluated by ${\beta _{j}}{\rm{ = }}\frac{{\left\langle {{e_{j}},{g_j+1}} \right\rangle }}{{{{\left\| {{g_j+1}} \right\|}^2}}}$. If the CFN-RW with $L - 1$ hidden nodes does not reach the termination condition, then a new hidden node will be generated by the following two steps:

1) The input weights ${\omega _L}$ and bias ${b_L}$ are randomly generated from the fixed interval ${\left[ { - \lambda ,\lambda } \right]^d}$ and $\left[ { - \lambda ,\lambda } \right]$. In particular, $\lambda $ usually takes the value 1. Then, the output vector ${g_L}$ of the $L$-th hidden node, which is determined by maximizing $\Delta  = \frac{{{{\left\langle {{e_{L - 1}},{g_L}} \right\rangle }^2}}}{{{{\left\| {{g_L}} \right\|}^2}}}$, where ${e_{L - 1}} = f - {f_{L - 1}} = \left[ {{e_{L - 1,1}},{e_{L - 1,2}},...,{e_{L - 1,m}}} \right]$ is the current network residual error.

2) The output weights vector ${\beta _L}$ of the $L$-th hidden node can be obtained by ${\beta _L}{\rm{ = }}\frac{{\left\langle {{e_{L - 1}},{g_L}} \right\rangle }}{{{{\left\| {{g_L}} \right\|}^2}}}$.

If the new network residual error ${e_L} = f - {f_L}$ dose not reach the predefined residual error, a new hidden node needs to be added until the predefined residual error or maximum number of hidden nodes is reached.
\subsection{CFN-RW Analysis}
According to ${e_{L,q}}{\rm{ = }}{e_{L - 1,q}} - {\beta _{L,q}}{g_L}$ and ${\beta _{L,q}}{\rm{ = }}\frac{{\left\langle {{e_{L - 1,q}},{g_L}} \right\rangle }}{{{{\left\| {{g_L}} \right\|}^2}}}$, we have
\begin{equation}
	\begin{array}{l}
		\quad{\rm{   }}\left\langle {{e_{L,q}},{g_L}} \right\rangle \\
		= \left\langle {{e_{L - 1,q}} - {\beta _{L,q}}{g_L},{g_L}} \right\rangle \\
		= \left\langle {{e_{L - 1,q}},{g_L}} \right\rangle  - {\beta _{L,q}}\left\langle {{g_L},{g_L}} \right\rangle \\
		{\rm{ = }}\left\langle {{e_{L - 1,q}},{g_L}} \right\rangle  - \frac{{\left\langle {{e_{L - 1,q}},{g_L}} \right\rangle }}{{{{\left\| {{g_L}} \right\|}^2}}}\left\langle {{g_L},{g_L}} \right\rangle \\
		= 0
	\end{array}
\end{equation}

Thus, Eq. (1) means ${e_{L,q}} \bot {g_L}$. Further, based on Eq. (1) and $\Delta  = \frac{{{{\left\langle {{e_{L - 1}},{g_L}} \right\rangle }^2}}}{{{{\left\| {{g_L}} \right\|}^2}}}$ when$L \to \infty $, we have

\begin{equation}
\begin{array}{l}
	\quad{\rm{  }}{\left\| {{e_L}} \right\|^2} - {\left\| {{e_{L - 1}}} \right\|^2}\\
	= {\left\| {{e_{L - 1}}} \right\|^2}{\sin ^2}{\theta _L} - {\left\| {{e_{L - 1}}} \right\|^2}\\
	= {\left\| {{e_{L - 1}}} \right\|^2}\left( {{{\sin }^2}{\theta _L} - 1} \right)\\
	\quad{\rm{  }} < 0
\end{array}
\end{equation}
where, ${\theta _L}{\rm{ = }}\angle \left( {{e_{L - 1}},{g_L}} \right)$. Then, we have

\begin{equation}
	\begin{array}{l}
		\frac{{{{\left\| {{e_{L{\rm{ - }}1}}} \right\|}^2} - {{\left\| {{e_L}} \right\|}^2}}}{{{{\left\| {{e_{L{\rm{ - }}1}}} \right\|}^2}}}{\rm{ = }}1{\rm{ - }}{\sin ^2}{\theta _L} < 1
\end{array}
\end{equation}

Let $1{\rm{ - }}{\sin ^2}{\theta _L}{\rm{ = }}{\varepsilon _L}$, we have
\begin{equation}
	\begin{array}{l}
		{\lim _{L \to \infty }}\prod\limits_{k = 1}^L {\left( {1{\rm{ - }}{\varepsilon _k}} \right)}  = \varepsilon  > 0
\end{array}
\end{equation}

According Eq. (3) and (4), we have
\begin{equation}
	\begin{array}{l}
	\quad{\rm{  }}{\left\| {{e_L}} \right\|^2}\\
	= \left( {1{\rm{ - }}{\varepsilon _L}} \right){\left\| {{e_{L{\rm{ - }}1}}} \right\|^2}\\
	{\rm{ = }}\prod\limits_{k = 1}^L {\left( {1{\rm{ - }}{\varepsilon _k}} \right){{\left\| f \right\|}^2}} 
\end{array}
\end{equation}

Then, we have
\begin{equation}
	\begin{array}{l}
		{\lim _{L \to \infty }}{\left\| {{e_L}} \right\|^2}{\rm{ = }}{\lim _{L \to \infty }}\prod\limits_{k = 1}^L {\left( {1{\rm{ - }}{\varepsilon _k}} \right){{\left\| f \right\|}^2}}  = \varepsilon {\left\| f \right\|^2}
\end{array}
\end{equation}

From the above analysis, we have
\begin{equation}
	\begin{array}{l}
		{\lim _{L \to \infty }}\left\| {f - {f_L}} \right\|{\rm{ = }}\sqrt \varepsilon  \left\| f \right\|
\end{array}
\end{equation}

The above analysis reveals that CFN-RW may not share the universal approximation property.

\section{Proposed Lightweight Geometric Constructive Neural Network}

This section establishes a compact angle constraint (CAC) based on the spatial geometric relationships to supervise the hidden parameters assignment process of LightGCNet. Then, we prove that a LightGCNet constructed by CAC has universal approximation ability. Moreover, a node pool strategy is employed in compact angle constraint to select hidden parameters that are more conducive to convergence. Finally, two different algorithm implementations are proposed, namely LightGCNet-I and LightGCNet-II.

\begin{figure}[!h]
	\centering
	\includegraphics[width=2.0in]{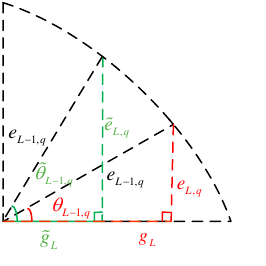}
	\caption{Spatial Geometric relationship between ${{e_L}}$, ${{e_{L - 1}}}$ and ${g_L}$.}
	\label{fig_1}
\end{figure}
\subsection{Compact Angle Constraint}
For a target function $f:{R^d} \to {R^m}$, assume that a LightGCNet-I with ${L-1}$ hidden nodes has been constructed, i.e., ${f_{L - 1}} = \sum\nolimits_{j = 1}^{L - 1} {{\beta _j}{g_j}\left( {\omega _j^{\rm{T}} \cdot x + {b_j}} \right)} $. If the residual error ${e_{L - 1}}$ does not meet the preset requirements, then ${g_L}$ satisfying compact angle constraint will be added to the hidden layer, and the output weight will be calculated by ${\beta _{L,q}}{\rm{ = }}\frac{{\left\langle {{e_{L - 1,q}},{g_L}} \right\rangle }}{{{{\left\| {{g_L}} \right\|}^2}}},q = 1,2,..,m$. Repeat the above steps until the residual error meets the preset requirements or the number of hidden layer nodes reaches the maximum. The main result of the LightGCNet can be stated in the following theorem.

\noindent \quad {\bf{Theorem 1:}} Suppose that span($\Gamma $) is dense in ${L^2}$ and $\forall g \in \Gamma $, $0 < \left\| g \right\| < v$ for some $v \in R$. For a given target function $f$,  if the newly added hidden node  ${g_L}$ meets the following compact angle constraint
\begin{equation}
\left( {{\theta _{L - 1,q}}} \right) \le \arccos \left( {\sqrt {\frac{\tau }{{{L^\mu } + \tau }}} \left\| {{e_{L - 1,q}}} \right\|} \right),{\rm{ }}q = 1,2,...,m.
\end{equation}
where ${{\theta _{L - 1,q}}}$ denotes the angle between the hidden node ${g_L}$ and the residual error ${e_{L - 1,q}}$, $\mu ,\tau  \in \left( {0,1} \right)$.

The output weights ${\beta _L} = \left[ {{\beta _{L,1}},{\beta _{L,2}}, \cdots ,{\beta _{L,m}}} \right]$ are evaluated by
\begin{equation}
 {\beta _{L,q}}{\rm{ = }}\frac{{\left\langle {{e_{L - 1,q}},{g_L}} \right\rangle }}{{{{\left\| {{g_L}} \right\|}^2}}}. 
\end{equation}
Then, we have ${\lim _{L \to  + \infty }}\left\| {{e_L}} \right\| = 0$.

The universal approximation property are fundamental aspects of the Theorem 1 for data modeling. Therefore, the proof is given below.

\noindent \quad {\bf{Proof:}}

Based on the CFN-RW, we have that
\begin{equation}
\begin{array}{l}
\quad {\rm{   }}{\left\| {{e_L}} \right\|^2} - {\left\| {{e_{L - 1}}} \right\|^2}\\
 = \sum\limits_{q = 1}^m {\left\langle {{e_{L - 1,q}} - {\beta _{L,q}}{g_L},{e_{L - 1,q}} - {\beta _{L,q}}{g_L}} \right\rangle } \\
{\rm{     }}\quad -\sum\limits_{q = 1}^m {\left\langle {{e_{L - 1,q}},{e_{L - 1,q}}} \right\rangle } \\
{\rm{ = }} - \sum\limits_{q = 1}^m {\frac{{{{\left\langle {{e_{L - 1,q}},{g_L}} \right\rangle }^2}}}{{{{\left\| {{g_L}} \right\|}^2}}}} \\
 \le 0
\end{array}
\end{equation}

Then, it has been proved that the residual error $\left\| {{e_L}} \right\|$ is monotonically decreasing as $L \to \infty $. Thus, Eq. (1) means ${e_{L,q}} \bot {g_L}$. It can be easily observed that ${{e_{L,q}}}$, ${{e_{L - 1,q}}}$ and ${g_L}$ satisfy the spatial geometric relationship shown in Fig. 1.

It follows from Eq. (9) and Eq. (10) that
\begin{equation}
\begin{array}{l}
\quad {\rm{  }}{\left\| {{e_L}} \right\|^2} - \left( {\frac{{{L^\mu }}}{{{L^\mu } + \tau }}} \right){\left\| {{e_{L - 1}}} \right\|^2} \\
 = \sum\limits_{q = 1}^m {\left\langle {{e_{L - 1,q}} - {\beta _{L,q}}{g_L},{e_{L - 1,q}} - {\beta _{L,q}}{g_L}} \right\rangle } \\
{\rm{     }} \quad - \sum\limits_{q = 1}^m {\left( {\frac{{{L^\mu }}}{{{L^\mu } + \tau }}} \right)\left\langle {{e_{L - 1,q}},{e_{L - 1,q}}} \right\rangle } \\
{\rm{ = }}\left( {1 - \left( {\frac{{{L^\mu }}}{{{L^\mu } + \tau }}} \right)} \right)\sum\limits_{q = 1}^m {{{\left\| {{e_{L - 1,q}}} \right\|}^2}}  - \sum\limits_{q = 1}^m {\frac{{{{\left\langle {{e_{L - 1,q}},{g_L}} \right\rangle }^2}}}{{{{\left\| {{g_L}} \right\|}^2}}}} \\
 \le \left( {\frac{\tau }{{{L^\mu } + \tau }}} \right)\sum\limits_{q = 1}^m {{{\left\| {{e_{L - 1,q}}} \right\|}^2}}  - \sum\limits_{q = 1}^m {\frac{{{{\left\langle {{e_{L - 1,q}},{g_L}} \right\rangle }^2}}}{{{{\left\| {{g_L}} \right\|}^2}}}} \\
 \le \left( {\frac{\tau }{{{L^\mu } + \tau }}} \right)\sum\limits_{q = 1}^m {{{\left\| {{e_{L - 1,q}}} \right\|}^2}}  - \sum\limits_{q = 1}^m {{{\cos }^2}{\theta _{L - 1,q}}} \\
 \le \sum\limits_{q = 1}^m {\left( {\left( {\frac{\tau }{{{L^\mu } + \tau }}} \right){{\left\| {{e_{L - 1,q}}} \right\|}^2} - {{\cos }^2}{\theta _{L - 1,q}}} \right)}
\end{array}
\end{equation}

In addition, based on ${f_{L - 1}} = \sum\limits_{j = 1}^{L - 1} {{\beta _j}{g_j}}$, we have
\begin{equation}
\begin{array}{l}
\quad {\rm{       }}\sum\limits_{j = 1}^{L - 1} {\sum\limits_{q = 1}^m {{\beta _{j,q}}\left\langle {{e_{L - 1,q}},{g_j}} \right\rangle } } \\
{\rm{   }} = \sum\limits_{q = 1}^m {\left\langle {{e_{L - 1,q}},\sum\limits_{j = 1}^{L - 1} {{\beta _{j,q}}{g_j}} } \right\rangle } \\
{\rm{   }} = \sum\limits_{q = 1}^m {\left\langle {{e_{L - 1,q}},{e_{L - 1,q}} + {f_{L - 1,q}}} \right\rangle } \\
{\rm{   }} = \sum\limits_{q = 1}^m {\left( {\left\langle {{e_{L - 1,q}},{e_{L - 1,q}}} \right\rangle  + \left\langle {{e_{L - 1,q}},{f_{L - 1,q}}} \right\rangle } \right)} \\
{\rm{   }} = \sum\limits_{q = 1}^m {\left\langle {{e_{L - 1,q}},{e_{L - 1,q}}} \right\rangle } \\
{\rm{   }} = {\left\| {{e_{L - 1}}} \right\|^2}
\end{array}
\end{equation}
where ${f_{L - 1,q}}$ is orthogonal to ${e_{L - 1,q}}$. Thus, we have
\begin{equation}
\begin{array}{l}
\qquad {\rm{       }}\exists {\beta _{j,q}}\left\langle {{e_{L - 1,q}},{g_j}} \right\rangle  \ge \frac{{{{\left\| {{e_{L - 1,q}}} \right\|}^2}}}{{L - 1}}\\
 <  =  > {\beta _{j,q}}\left\| {{g_j}} \right\|\frac{{\left\langle {{e_{L - 1,q}},{g_j}} \right\rangle }}{{\left\| {{g_j}} \right\|}} \ge \frac{{{{\left\| {{e_{L - 1,q}}} \right\|}^2}}}{{L - 1}}\\
{\rm{ }} =  > \frac{{\left| {\left\langle {{e_{L - 1,q}},{g_j}} \right\rangle } \right|}}{{\left\| {{g_j}} \right\|}} \ge \frac{{{{\left\| {{e_{L - 1,q}}} \right\|}^2}}}{{\left( {L - 1} \right){\beta _{j,q}}\left\| {{g_j}} \right\|}}
\end{array}
\end{equation}
where $\frac{{{{\left\| {{e_{L - 1,q}}} \right\|}^2}}}{{L - 1}}$ is the average of ${{{\left\| {{e_{L - 1,q}}} \right\|}^2}}$.

According to the spatial geometric relationship (Fig. 1) and Eq. (13), the following equation is obtained
\begin{equation}
\frac{{\left| {\left\langle {{g_j},{e_{L - 1,q}}} \right\rangle } \right|}}{{\left\| {{g_j}} \right\|}} = \left\| {{e_{L - 1,q}}} \right\|{\cos}{\theta _{L - 1,q}}
\end{equation}

It follows from Eq. (13) and (14) that
\begin{equation}
{\cos ^2}{\theta _{L - 1,q}} \ge \varphi {\left\| {{e_{L - 1,q}}} \right\|^2}
\end{equation}
where $\varphi  = \frac{1}{{{{\left( {\left( {L - 1} \right){\beta _{j,q}}\left\| {{g_j}} \right\|} \right)}^2}}}$, which is sufficiently small when $L - 1$ is very large.

According to the inverse trigonometric functions, Eq. (15) can be rewritten as
\begin{equation}
{\theta _{L - 1,q}} \le \arccos \left( {\sqrt \varphi  \left\| {{e_{L - 1,q}}} \right\|} \right)
\end{equation}

Eq. (16) shows that the random assignment of node parameters in the incremental process should satisfy some constraints. In addition, as the modeling process proceeds, the residual error becomes smaller which makes the configuration task on ${w_L}$  and ${b_L}$  more challenging. Therefore, the parameter ${{\gamma _L} = \frac{\tau }{{{L^\mu } + \tau }}}$ is designed as a dynamic value to ensure that Eq. (17) holds
\begin{equation}
\varphi  \ge \frac{\tau }{{{L^\mu } + \tau }}
\end{equation}
where $0 < \tau ,\mu  < 1$, $L$ denotes the number of hidden nodes.

It follows from Eq. (8), (11), (16), and (17) that
\begin{equation}
\begin{array}{l}
\quad {\rm{   }}{\left\| {{e_L}} \right\|^2} - \left( {\frac{{{L^\mu }}}{{{L^\mu } + \tau }}} \right){\left\| {{e_{L - 1}}} \right\|^2}\\
 \le \sum\limits_{q = 1}^m {\left( {\left( {\frac{\tau }{{{L^\mu } + \tau }}} \right){{\left\| {{e_{L - 1,q}}} \right\|}^2} - {{\cos }^2}{\theta _{L - 1,q}}} \right)} \\
 \le 0
\end{array}
\end{equation}

Besides, when $L \to \infty $
\begin{equation}
\small
{\lim _{L \to  + \infty }}{\left\| {{e_L}} \right\|^2} \le \left( {\frac{{{L^\mu }}}{{{L^\mu } + \tau }}} \right){\left\| {{e_{L - 1}}} \right\|^2} \le  \cdots  \le \prod\limits_{i = 1}^L {\left( {\frac{{{i^\mu }}}{{{i^\mu } + \tau }}} \right){{\left\| f \right\|}^2}}
\end{equation}
and
\begin{equation}
{\lim _{L \to  + \infty }}\prod\limits_{i = 1}^L {\left( {\frac{{{i^\mu }}}{{{i^\mu } + \tau }}} \right)}  = 0
\end{equation}

Then, we have that ${\lim _{L \to  + \infty }}\left\| {{e_L}} \right\| = 0$. $\hfill\qedsymbol$

\subsection{Node Pool Strategy}
In LightGCNet, the hidden parameters (${g_L}$) are assigned randomly at a single time, which may not make the network residual decrease quickly. As a result, Eq. (1) is optimized using the node pool strategy as
\begin{equation}
{\left( {{\theta _{L - 1,q}}} \right)_{\min }} \le \arccos \left( {\sqrt {\frac{\tau }{{{L^\mu } + \tau }}} \left\| {{e_{L - 1,q}}} \right\|} \right),{\rm{ }}q = 1,2,...,m.
\end{equation}

In node pool strategy, several hidden parameters are randomly generated, and the one leading to the minimum $\theta _{L - 1,q}$ will be added to the existing  LightGCNet. The construction process of LightGCNet is shown in Fig. 2.
\begin{figure}[!h]
	\centering
	\includegraphics[width=3.6in]{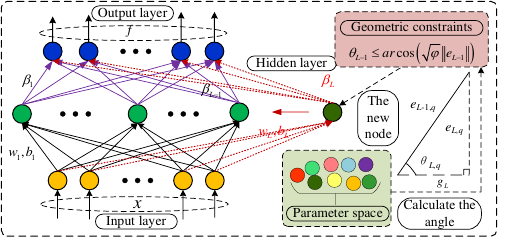}
	\caption{Construction Process of LightGCNet.}
	\label{fig_2}
\end{figure}

\subsection{Algorithm Implementations}
Two algorithm implementations of Theorem 1 are shown in this section, namely LightGCNet-I and LightGCNet-II. 

\subsubsection{LightGCNet-I}
For a target function $f:{R^d} \to {R^m}$, assume that a LightGCNet-I with ${L-1}$ hidden nodes has been constructed, i.e., ${f_{L - 1}} = \sum\nolimits_{j = 1}^{L - 1} {{\beta _j}{g_j}\left( {\omega _j^{\rm{T}} \cdot x + {b_j}} \right)} $. If the residual error ${e_{L - 1}}$ does not meet the preset requirements, then ${g_L}$ satisfying Eq. (15) will be added to the hidden layer, and the output weight will be calculated by ${\beta _{L,q}}{\rm{ = }}\frac{{\left\langle {{e_{L - 1,q}},{g_L}} \right\rangle }}{{{{\left\| {{g_L}} \right\|}^2}}},q = 1,2,..,m$. Repeat the above steps until the residual error meets the preset requirements or the number of hidden layer nodes reaches the maximum.

\subsubsection{LightGCNet-II}
The traditional output weights calculation method (${\beta _L}{\rm{ = }}\frac{{\left\langle {{e_{L - 1}},{g_L}} \right\rangle }}{{{{\left\| {{g_L}} \right\|}^2}}}$) may lead to slower convergence. Therefore, a global optimization method is designed to evaluate the output weights, i.e., LightGCNet-II. For a target function $f:{R^d} \to {R^m}$, assume that a LightGCNet-II with ${L-1}$ hidden nodes has been constructed, i.e., ${f_{L - 1}} = \sum\nolimits_{j = 1}^{L - 1} {{\beta _j}{g_j}\left( {\omega _j^{\rm{T}} \cdot x + {b_j}} \right)} $. If the residual error ${e_{L - 1}}$ does not meet the preset requirements, then ${g_L}$ satisfying Eq. (21) will be added to the hidden layer, and the output weights will be calculated by
\begin{equation}
\beta {\rm{ = }}\arg \mathop {\min }\limits_\beta  \left\| {f - \sum\limits_{j = 1}^L {{\beta _j}{g_j}} } \right\|
\end{equation}

Rearranging Eq.(16), the following matrix form results
\begin{equation}
\label{deqn_ex1a}
\beta  = {H^\dag }{f_L}
\end{equation}
where $\beta  = \left[ {{\beta _1},{\beta _2},...,{\beta _L}} \right]_{}^{\rm{T}}$,  $H = \left[ {{g_1},{g_2},...,{g_L}} \right]$, ${H^\dag }$ denotes the Moore-Penrose generalized inverse of $H$.

Then, we have that ${\lim _{L \to \infty }}\left\| {f - {f_L}} \right\|{\rm{ = 0}}$.

\noindent \quad {\bf{Proof:}}

Let ${\beta ^ * } = \left[ {\beta _1^ * ,\beta _2^ * , \cdots ,\beta _L^ * } \right] = \arg {\min _\beta }\left\| {f - \sum\nolimits_{j = 1}^L {{\beta _j}{g_j}} } \right\|$, $e_L^ *  = f - \sum\nolimits_{j = 1}^L {\beta _j^ * {g_j}} $. Define the intermediate values ${\hat \beta _L} = {{\left\langle {e_{L - 1}^ * ,{g_L}} \right\rangle } \mathord{\left/
 {\vphantom {{\left\langle {e_{L - 1}^ * ,{g_L}} \right\rangle } {{{\left\| {{g_L}} \right\|}^2}}}} \right.
 \kern-\nulldelimiterspace} {{{\left\| {{g_L}} \right\|}^2}}}$ and ${\hat e_L} = e_{L - 1}^ *  - {\hat \beta _L}{g_L}$, $e_{}^ *  = f$. Then, we have ${\left\| {e_L^ * } \right\|^2} \le {\left\| {{{\hat e}_L}} \right\|^2} = {\left\| {e_{L - 1}^ *  - {{\hat \beta }_L}{g_L}} \right\|^2} \le {\left\| {e_{L - 1}^ * } \right\|^2} \le {\left\| {{{\hat e}_{L - 1}}} \right\|^2}$ for $L \ge 2$, thus $\left\{ {\left\| {e_L^ * } \right\|} \right\}$  is monotonically decreasing sequence and convergent. Similar to the proof of Theorem 1, we have ${\left\| {e_L^ * } \right\|^2} - \left( {\frac{{{L^\mu }}}{{{L^\mu } + \tau }}} \right){\left\| {e_L^ * } \right\|^2} \le 0$.

Then, based on Eq. (19) and (20), we have that ${\lim _{L \to  + \infty }}\left\| {e_L^ * } \right\| = 0$. $\hfill\qedsymbol$

\section{Case Study}
In this section, we present the performance of the proposed LightGCNet-I and LightGCNet-II as well as CFN-RW, and TIC on a function approximation dataset, five benchmark datasets\cite{ref19,ref20}, an ore grinding process\cite{ref21}. The function approximation dataset was randomly generated by Eq. (24) defined on [0, 1]. The specifications of the six datasets can be found in TABLE I. In addition, the experimental parameters of all algorithms are summarized in TABLE II. The above experimental parameter settings are the best solutions obtained from multiple experiments. All the comparing experiments are implemented on MATLAB 2020a running on a PC with 3.00 GHz Core i7 CPU and 8 GB RAM. Each experiment is repeated 30 times, and the average of the 30 experiments is set as the final reported result. The sigmoid function g($x$) = 1/(1 + exp(${-x}$)) is employed as the activation function of these four randomized algorithms. In addition, root mean squares error (RMSE) and efficiency (the time spent on building the network) are employed to measure the performance of all randomized algorithms.
\begin{equation}
f\left( x \right) = \frac{1}{{({{(x - 0.3)}^2} + 0.01)}} + \frac{1}{{({{(x - 0.9)}^2} + 0.04)}} - 6
\end{equation}
where $x \in \left[ {0,1} \right]$.
\begin{table}[!t]
\caption{Description of Five Benchmark Datasets\label{tab:table1}}
\centering
\begin{tabular}{ccccc}
\hline
Datasets & Training &Testing & Features & Output\\
\hline
Function& 2000 & 400 & 1 & 1\\
Winequality & 1120 & 479 & 11 & 1\\
Anacal & 2836 & 1216 & 7 & 1\\
Delta\_ail & 4990 & 2139 & 5 & 1\\
Plastic & 1155 & 496 & 2 & 1\\
Compactiv & 6144 & 2048 & 21 & 1\\
\hline
\end{tabular}
\end{table}
\begin{table}[!t]
\caption{Experimental Parameters\label{tab:table2}}
\centering
\begin{tabular}{ccccccc}
\hline
 \multirow{2}{*}{Datasets} & \multicolumn{2}{c}{ CFN-RW}& \multicolumn{2}{c}{TIC/LightGCNet} &  \multirow{2}{*}{$\ell $}  & \multirow{2}{*}{${L_{\max }}$}  \\
 \cline{2-5}
  &  $\lambda $ & ${T_{\max }}$ & $\zeta $& ${T_{\max }}$ & &\\
\hline
Function & 150& & 150:10:200 &   & 0.05 & 200\\
Winequality & 0.5& &0.5:0.1:5 & & 0.05 & 100\\
Anacal & 1& & 0.5:0.1:5 &   & 0.05 & 150\\
Delta\_ail & 1&1 & 1:10:100 &  20 & 0.05 & 100\\
Plastic & 0.5& & 0.5:10:200 &   & 0.05 & 100\\
Compactiv & 0.5& & 1:10:50 &   & 0.05 & 100\\
\hline
\end{tabular}
\end{table}
\subsection{Benchmark Case}
\subsubsection{Function Approximation Dataset}
To investigate the advantages of the proposed algorithms, this part compares LightGCNet-I, LightGCNet-II, CFN-RW, and TIC on the function approximation dataset. Fig. 3 and Fig. 4 list the experimental results of RSME, node utilization, and time for the four algorithms, respectively. Node utilization indicates the usage of the pre-defined nodes by the algorithm. By observing the two figures, we can see that (1) Under the constraint of Eq. (21), the RMSE of LightGCNet-I and LightGCNet-II stops at 0.1 and 0.0008, respectively. This result indicates that the prediction performance of LightGCNet is stable and good. (2) As for the proposed LightGCNet-I, the RMSE is higher than TIC but lower than CFN-RW. This indicates that the LightGCNet-I performs better than CFN-RW, but worse than TIC in terms of generalization. (3) LightGCNet-II and TIC achieve similar RMSE, but LightGCNet-II is more stable. (4) The node utilization of both CFN-RW and LightGCNet-I is 1, while LightGCNet-I is more time consuming because it uses a node pool strategy to select hidden parameters. This result indicates that the proposed Eq. (15) increases the modeling time. (5) Compared with TIC, LightGCNet-II has a significant advantage in terms of node utilization and time. This indicates that the proposed LightGCNet-II is superior to TIC in network size and fast modeling. In addition, the experimental results of LightGCNet-I and LightGCNet-II also prove the deficiency of Eq. (16).
\begin{figure}[!t]
\centering
\includegraphics[width=3.2in]{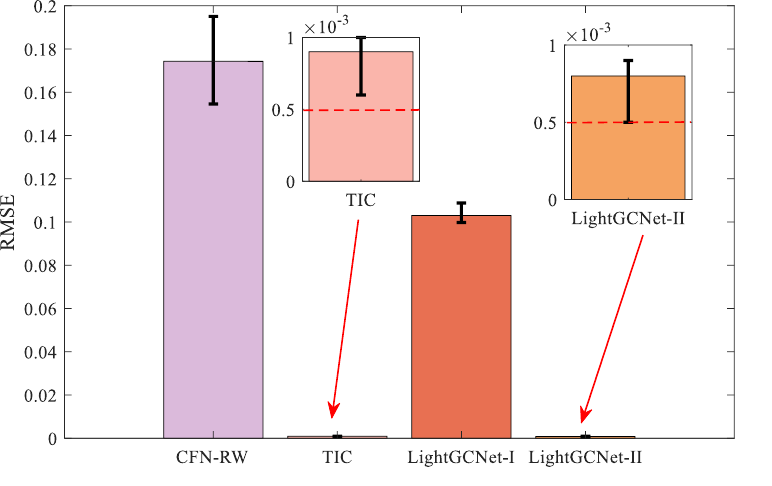}
\caption{RMSE of the CFN-RW, TIC, LightGCNet-I, and LightGCNet-II.}
\label{fig_3}
\end{figure}
\begin{figure}[!t]
\centering
\includegraphics[width=3.3in]{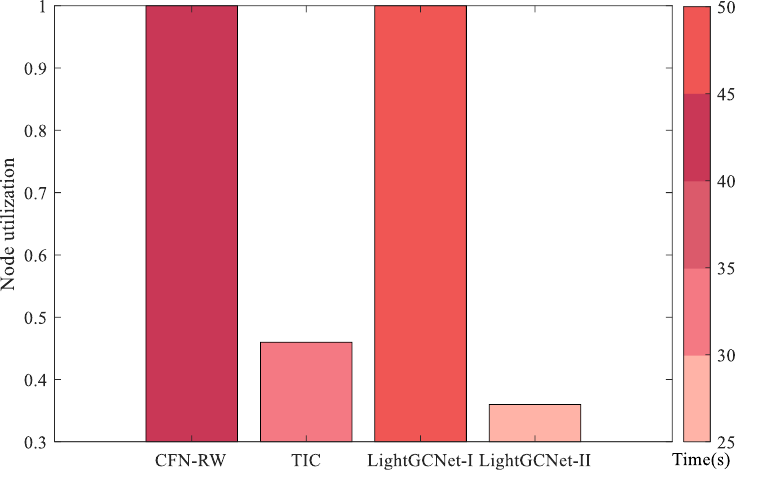}
\caption{Node utilization and time results for the CFN-RW, TIC, LightGCNet-I, and LightGCNet-II.}
\label{fig_4}
\end{figure}

\begin{table}[!t]
	\caption{Comparison results of hidden parameter assignment and output weight evaluation of four algorithms\label{tab:table3}}
	\centering
	\begin{tabular}{ccccc}
		\hline
		\multirow{2}{*}{Algorithms}& \multicolumn{2}{c}{Hidden parameter}& \multicolumn{2}{c}{Output weights}\\
		\cline{2-5}
		&Random&Constraint&Eq. (9)& Eq. (23)\\
	\hline
	CFN-RW&\checkmark&&\checkmark& \\
	TIC& \checkmark&\checkmark&&\checkmark\\
	LightGCNet-I& \checkmark&\checkmark&\checkmark&\\
	LightGCNet-II&\checkmark &\checkmark&&\checkmark\\
		\hline
	\end{tabular}
\end{table}

\begin{figure}[!t]
	\centering
	\includegraphics[width=3.3in]{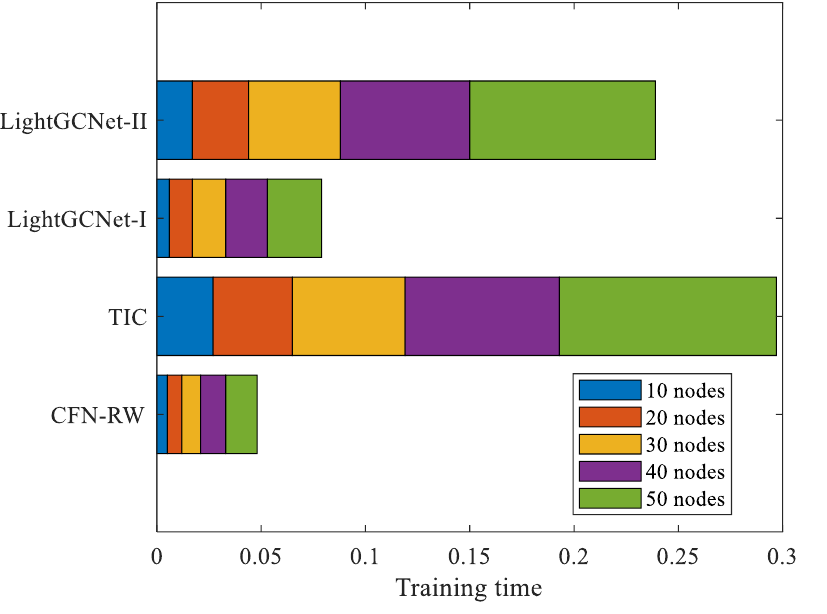}
	\caption{Modeling time under different nodes.}
	\label{fig_5}
\end{figure}

\begin{table}[!t]
	\caption{Number of nodes in different RMSE.\label{tab:table5}}
	\centering
	\begin{tabular}{cccccc}
		\hline
		\multirow{2}{*}{Algorithms} & \multicolumn{5}{c}{RMSE}   \\
		\cline{2-6}
		& 0.01 &0.02 & 0.03&0.04&0.1\\
		\hline
		CFN-RW& - & - & - &-&-\\
		TIC& 45.21 & 36.33 & 28.00 &18.00&10.67\\
		LightGCNet-I& - & - & - &-&173.33\\
		LightGCNet-II & \textbf{41.33 }& \textbf{32.33}& \textbf{23.33}&\textbf{15.00}&\textbf{9.33}\\
		\hline
	\end{tabular}
\end{table}
\begin{table}[!t]
	\caption{Impact of node pool on RMSE.\label{tab:table6}}
	\centering
	\begin{tabular}{ccc}
		\hline
\multirow{2}{*}{$T_{\max } $}& \multicolumn{2}{c}{Algorithms}\\
\cline{2-3}
&LightGCNet-I&LightGCNet-II\\
\hline
10&0.2084±0.0262 &\textbf{0.0011±1.7520e-04} \\
20&0.1852±0.0277 & \textbf{0.0011±1.5333e-04}\\
30&0.2381±0.0281 & \textbf{0.0004± 1.6604e-04}\\
50&0.2110±0.0216 & \textbf{0.0008±1.9108e-04}\\
	\hline
	\end{tabular}
	\end{table}

In order to facilitate further analysis of the experimental results of each algorithm, this article compares and analyzes  all algorithms from two aspects: hidden parameters assignment and evaluation method of output weights, as shown in TABLE III. In order to test lightweight of the proposed LightGCNet in terms of modeling time and network size, this article studied the changes in modeling time of all algorithms when the number of hidden nodes was gradually increased, as shown in Fig. 5. Without loss of generality, CFN-RW always maintains the shortest modeling time due to its hidden parameters are generated by random assignment without constraints, and the output weights are obtained by stepwise calculation with low computational complexity. As can be seen from Fig. 5, with the increase in number of hidden nodes, the modeling time of LightGCNet-I and LightGCNet-II is always lower than that of TIC, especially LightGCNet-I. LightGCNet-II and TIC are inconsistent only in the hidden parameters assignment. As can be seen from Fig. 5, with the increase in a number of hidden nodes, the modeling time of LightGCNet-I and LightGCNet-II is always lower than that of TIC, especially LightGCNet-I. LightGCNet-II and TIC are inconsistent only in the hidden parameters assignment. With the same number of hidden nodes, LightGCNet-II requires less modeling time, which indicates that the constraint method proposed in this article has advantage of lightweight in terms of modeling time. In addition, this article studied the consumption of hidden node numbers of all algorithms under the same RMSE, as shown in TABLE IV ("-" indicates a null value). In particular, CFN-RW  is unable to achieve the preset RMSE. However, the existence of compact angle constraints enables LightGCNet-I to achieve 0.1 RMSE, which indicates that the proposed method can improve network performance. In addition, it can be seen that with the gradual increase of RMSE, the number of modeling nodes of LightGCNet-II is significantly lower than that of TIC, that is, LightGCNet-II can achieve the same RMSE with a smaller network structure. In summary, the algorithm proposed in this article is good lightweight in terms of modeling time and network size.

To explore the effect of node pool ${T_{\max }}$ on network performance, this article conducts an experimental study using LightGCNet-I and LightGCNet-II on the function approximation dataset, and the experimental results are shown in TABLE V.  It shows that too large or too small ${T_{\max }}$  leads to an increase in RMSE. It is worth noting that we do not show the time. In fact, ${T_{\max }}$ is  related to time because it controls the number of hidden parameters obtained from the random interval. In our experiments, this parameter was set with careful trade-offs.

\subsubsection{Benchmark Datasets}
In this section, the performance of the LightGCNet-I, LightGCNet-II, CFN-RW, and TIC is measured on five benchmark datasets. These benchmark datasets are mainly from KEEL and UCI, and their details can be observed in TABLE I. For each dataset, 70\% of them are randomly selected for training and the remaining for testing. TABLE VI shows the detailed experimental performances of the training time, training RMSE, and testing RMSE in CFN-RW, TIC, LightGCNet-I, and LightGCNet-II under the same experimental conditions. It can be seen from TABLE VI that the LightGCNet-II achieves the lowest or near-lowest in RMSE in both training and testing data on various datasets. This means that the LightGCNet-II outperforms other algorithms in terms of learning efficiency and generalization. Furthermore, the LightGCNet-I has a lower RSME than CFN-RW but a higher than TIC. This demonstrates the efficiency with which the proposed compact angle constraint improves generalization and learning efficiency. At the same time, the above-mentioned results show that the method for evaluating output weights is critical for the random algorithms. In addition, the LightGCNet-II takes less time to model than TIC with the same number of hidden nodes. This means that the compact angle constraint proposed in this article is lightweight. Therefore, the proposed LightGCNet-II has a more accurate and stable modeling performance.
\begin{table*}[!t]
    \caption{Comparison with Four Algorithms in Terms of Time, Training RMSE and Testing RMSE on the benchmark datasets.}
    \centering
\begin{tabular}{ccccccccccccc}
        \hline
\multirow{2}{*}{Dataset} & \multicolumn{12}{c}{Training time Training RMSE Testing RMSE}   \\
                         \cline{2-13}
                         & \multicolumn{3}{c}{CFN-RW} & \multicolumn{3}{c}{TIC} & \multicolumn{3}{c}{LightGCNet-II}  & \multicolumn{3}{c}{LightGCNet-I}\\
                         \hline
Winequality-read      & \textbf{0.120}s   & 0.308  & 0.308  & 0.314s   & \textbf{0.150}  & 0.159 & 0.287s   & \textbf{0.150} & \textbf{0.155 }& 0.169s   & 0.250 & 0.285\\
Anacal                 & \textbf{0.318}s  &0.366 & 0.376  & 1.092s   & 0.225  & 0.234 & 1.254s  &\textbf{0.280}& \textbf{0.284} & 0.396s   & 0.397 & 0.409\\
Delta\_ail            & \textbf{0.229}s   & 0.139 & 0.149  & 1.132s  & 0.082  & 0.085 & 1.104s & \textbf{0.078} &\textbf{0.082} & 1.131s  & 0.082& 0.085\\
Plastic                  & \textbf{0.152}s   & 0.551  & 0.559  & 0.506s   & 0.300  & 0.309 & 0.474s   & \textbf{0.288} & \textbf{0.297} & 0.230s  & 0.339 & 0.348\\
Compactiv                  & \textbf{0.272}s   & 0.756  & 0.760  & 2.007s   & 0.095  & 0.106 & 1.532s   & \textbf{0.084} & \textbf{0.100} & 0.467s  & 0.683 & 0.688\\
        \hline
\end{tabular}
\end{table*}
\subsection{Ore Grinding Case}
\begin{table}[!t]
	\caption{Description of Ore Griding Process Variables\label{tab:table1}}
	\centering
	\begin{tabular}{cc}
		\hline
		Variables & Descriptions\\
		\hline
		${R_1}$ & Fresh ore feed rate\\
		${R_2}$ & Mile inlet water flow rate\\
		${R_3}$ & Classifier overflow concentration\\
		${\alpha _1}$ & Current through mill\\
		${\alpha _2}$ & Current through classifier\\
		\hline
	\end{tabular}
\end{table}
\begin{figure}[!b]
	\centering
	\includegraphics[width=3.3in]{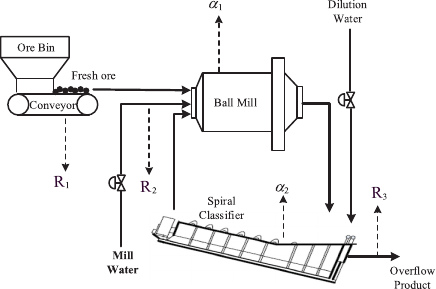}
	\caption{Flow chart of an ore grinding.}
	\label{fig_5}
\end{figure}
\begin{figure}[!t]
	\centering
	\includegraphics[width=3.3in]{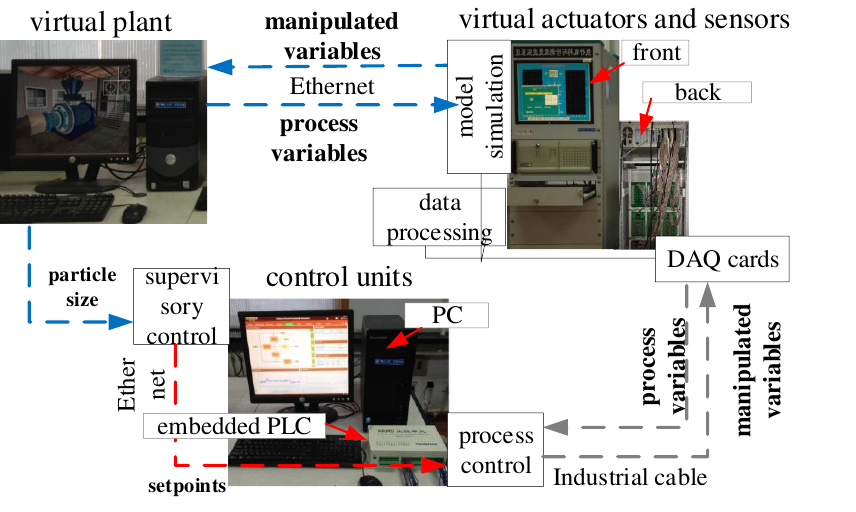}
	\caption{Structure of the Semi-physical Simulation Platform.}
	\label{fig_6}
\end{figure}
Ore grinding is the monomer dissociation between useful minerals and lode minerals, and its process is illustrated in Fig. 6. The ore grinding process is characterized by high capital investment, high electricity and material consumption, and the significant impact of mill capacity and grinding product size on the techno-economic indicators of efficiency and quality of subsequent operations. Therefore, the ore grinding process has been widely concerned and highly valued at home and abroad.In fact, the mechanisms of ore grinding are complicated and hard to establish a mathematical model. Therefore, it is essential to establish a soft sensors model for monitoring the ore grinding.

In the actual ore grinding process, particle size (PS) is usually used as an important process index to evaluate the quality of grinding products. Particle size refers to the proportion of particles smaller than 0.074 mm in diameter in ore products. Typically, if the particle size is too small, the useful minerals after dissociation are excessively broken, which is difficult to recycle and increase energy consumption. If the particle size is too large, it is difficult to achieve monomer dissociation of useful minerals. Therefore, it is important to estimate the ore grinding particle size. According to Fig. 6 and TABLE VII, the process variables affecting grinding particle size mainly include ${{R_1}}$, ${{R_2}}$, ${{R_3}}$, ${{\alpha _1}}$ and ${{\alpha _2}}$. In this article, these process variables are collected from the ore grinding semi-physical simulation platform (see Fig. 7) and 20000 training samples and 5000 test samples have been obtained. The purpose of constructing the soft sensors model of the ore grinding process is to achieve the following nonlinear mapping:
\begin{equation}
{\rm{PS}} = f\left( {{R_1},{R_2},{R_3},{\alpha _1},{\alpha _2}} \right)
\end{equation}
\subsubsection{Parameter Configuration}
For CFN-RW, the random set of hidden parameters is fixed interval [-150,150]. The random parameters of the other three randomized algorithms are selected from a variable interval $\zeta {\rm{ = }}\left\{ {150:1:500} \right\}$. For LightGCNet-I, LightGCNet-II, and TIC, the maximum number of iteration is set to ${L_{\max }}$ = 100, and the maximum times of random configuration is set to ${T_{\max }}$ = 20.
\subsubsection{Comparison and Discussion}
\begin{figure}[!t]
	\centering
	\includegraphics[width=3in]{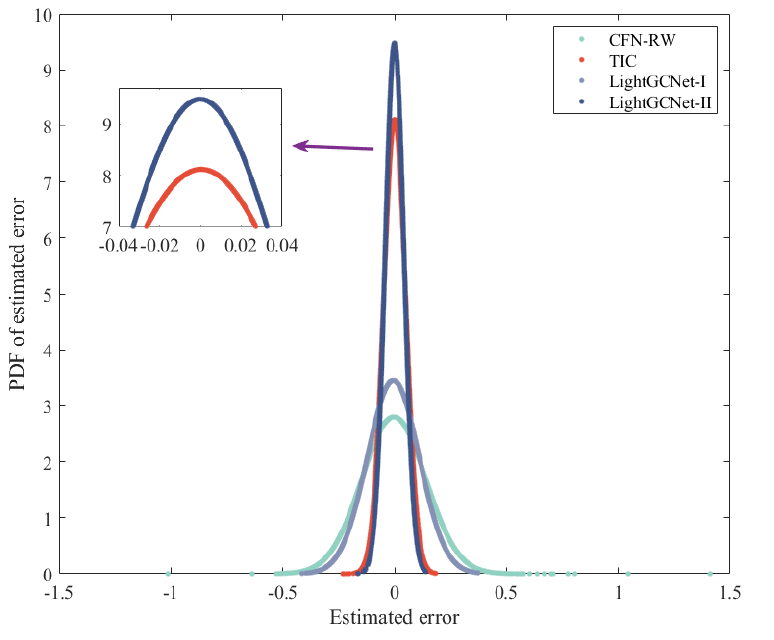}
	\caption{PDF of the CFN-RW, TIC, LightGCNet-I, and LightGCNet-II.}
	\label{fig_7}
\end{figure}
\begin{figure}[!t]
	\centering
	\includegraphics[width=3.3in]{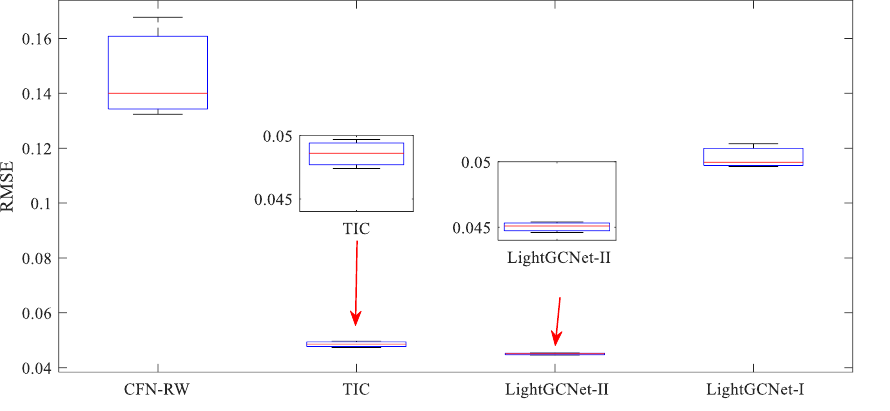}
	\caption{ RMSE box plot of the CFN-RW, TIC, LightGCNet-I, and LightGCNet-II.}
	\label{fig_8}
\end{figure}
Fig. 8 shows the probability density function (PDF) of the estimation error when the model reaches the expected error tolerance for the four models. It can be observed in Fig. 8 that the error PDF distribution curve of the proposed LightGCNet-II shows a Gaussian distribution shape that is both fine and high, and overall coincides with the vertical axis of "0". These results indicate that the estimation error of the proposed LightGCNet-II has a mean value of 0 in a probabilistic sense, i.e., the error between the actual and estimated values of the grinding model built using the proposed LightGCNet-II is relatively small. This means that the LightGCNet-II has the best performance among ore grinding systems. Besides, Fig. 9 shows the RMSE of the CFN-RW, LightGCNet-I, LightGCNet-II, and TIC on the ore grinding semi-physical simulation platform dataset. It can be seen from Fig. 9 that the box plot of LightGCNet-II is narrower compared with the other three randomized algorithms, which indicates that its RMSE is more stable. Meanwhile, the median RSME of LightGCNet-II is distributed around 0.045, which indicates that the LightGCNet-II has good generalization performance. In addition, the RMSE of the proposed LightGCNet-I is stable around 0.114 and lower than that of CFN-RW, which indicates that the compact angle constraint can improve the RMSE of the network. The overall box plot of CFN-RW is wide, which indicates that its RMSE is not stable. At the same time, the RMSE of TIC is slightly higher than that of LightGCNet-II, which further proves the effectiveness of the compact angle constraint proposed in this article.

\begin{figure*}[!t]
	\centering
	\includegraphics[width=5.6in]{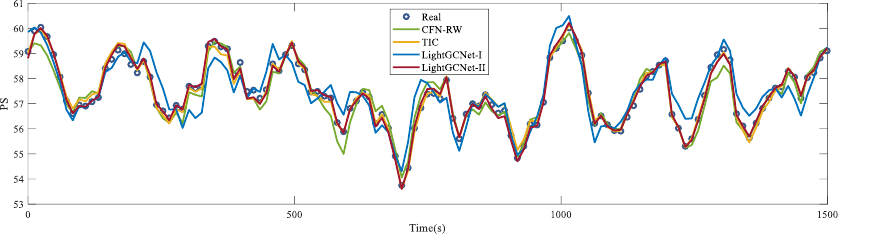}
	\includegraphics[width=5.6in]{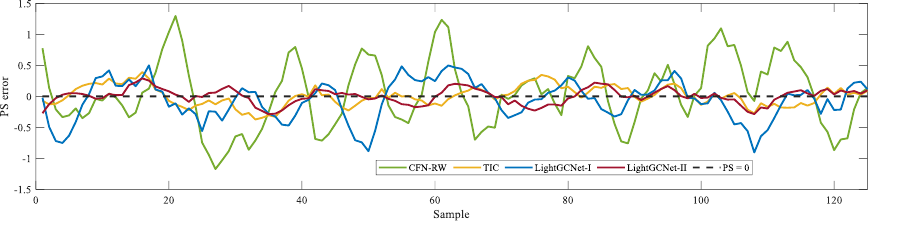}
	\caption{Modeling results of the CFN-RW, LightGCNet-I, LightGCNet-II, and TIC.}
	\label{fig_9}
\end{figure*}
\begin{figure}[!t]
\centering
\includegraphics[width=3in]{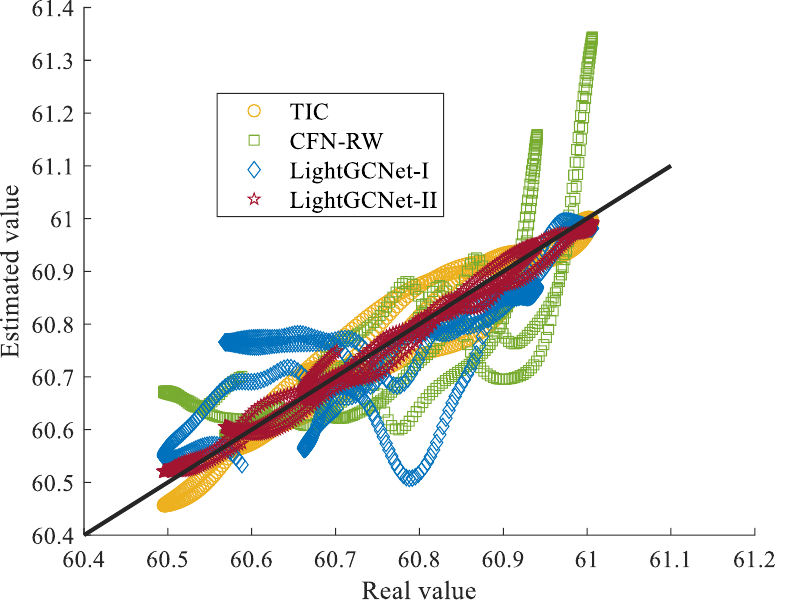}
\caption{Scatter plot of the CFN-RW, LightGCNet-I, LightGCNet-II, and TIC.}
\label{fig_10}
\end{figure}
Further, the actual data (sampling interval is 14s) of the beneficiation plant of Fushun Hanwang Maogong were used to test the proposed algorithm. The experimental results are shown in Fig. 10 and Fig. 11. It can be seen from Fig. 11 that the proposed LightGCNet-II has the best estimated performance for grinding particle size, and its estimated trend is generally consistent with the actual data. In addition, the PS error fluctuation range of LightGCNet-II is the smallest, which indicates that the proposed algorithm is more stable and has better generalization In the actual ore grinding process. As can be seen from Fig. 10, the autoregressive scatter plot of the proposed LightGCNet-II shows a very narrow distribution shape and an overall fit to the autoregressive line. This indicates that the error between the prediction results of the proposed LightGCNet-II and the actual values is very low, i.e., the ore grinding model established by LightGCNet-II is able to predict the ore grinding particle size better. In summary, the proposed method in this article is a promising choice to solve the prediction problem of ore grinding particle size.
\section{Conclusion}
In this work, we introduced a lightweight geometric constructive neural network, called LightGCNet, which is designed for constructing a data-driven soft sensors model on resource-constrained DCS devices. LightGCNet incorporates a compact angle constraint to assign the hidden parameter, ensuring convergence. The assignment process of hidden parameters is visualized and optimized using a node pool strategy and spatial geometric relationships. Moreover, we propose two different algorithm implementations, namely LightGCNet-I and LightGCNet-II. To demonstrate the superior performance of LightGCNet, we employ five benchmark datasets, assessing modeling speed, RMSE, and network size. Our experimental results in the context of ore grinding processes illustrate the efficacy of the proposed LightGCNet soft sensors model in predicting ore grinding particle size.

In future work, we help to develop LightGCNet with continuous learning capabilities to handle both class incremental environments and dynamic data changes in modeling tasks.

\begin{IEEEbiography}[{\includegraphics[width=1in,height=1.25in,clip,keepaspectratio]{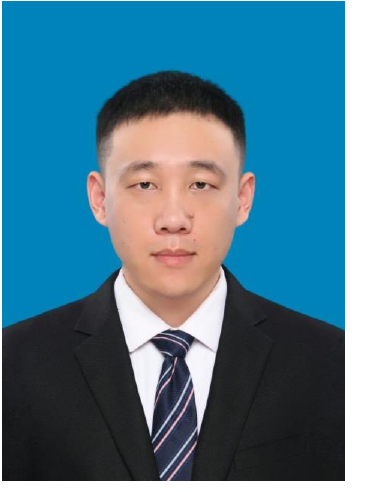}}]{Jing Nan }
	(Student Member, IEEE) received the B.S. degree in mathematics and applied mathematics from the school of science, Liaoning Petrochemical University, Fushun, China, in 2016, and the M.S. degree in control science and engineering in 2021 from the School of Information and Control Engineering, China University of Mining and Technology, Xuzhou, China, where he is currently working toward the Ph.D. degree in control science and engineering. His research interests include data-driven process monitoring, soft sensor modeling, and random weight neural networks.	
\end{IEEEbiography}
\vspace{-1.5cm}
\begin{IEEEbiography}[{\includegraphics[width=1in,height=1.25in,clip,keepaspectratio]{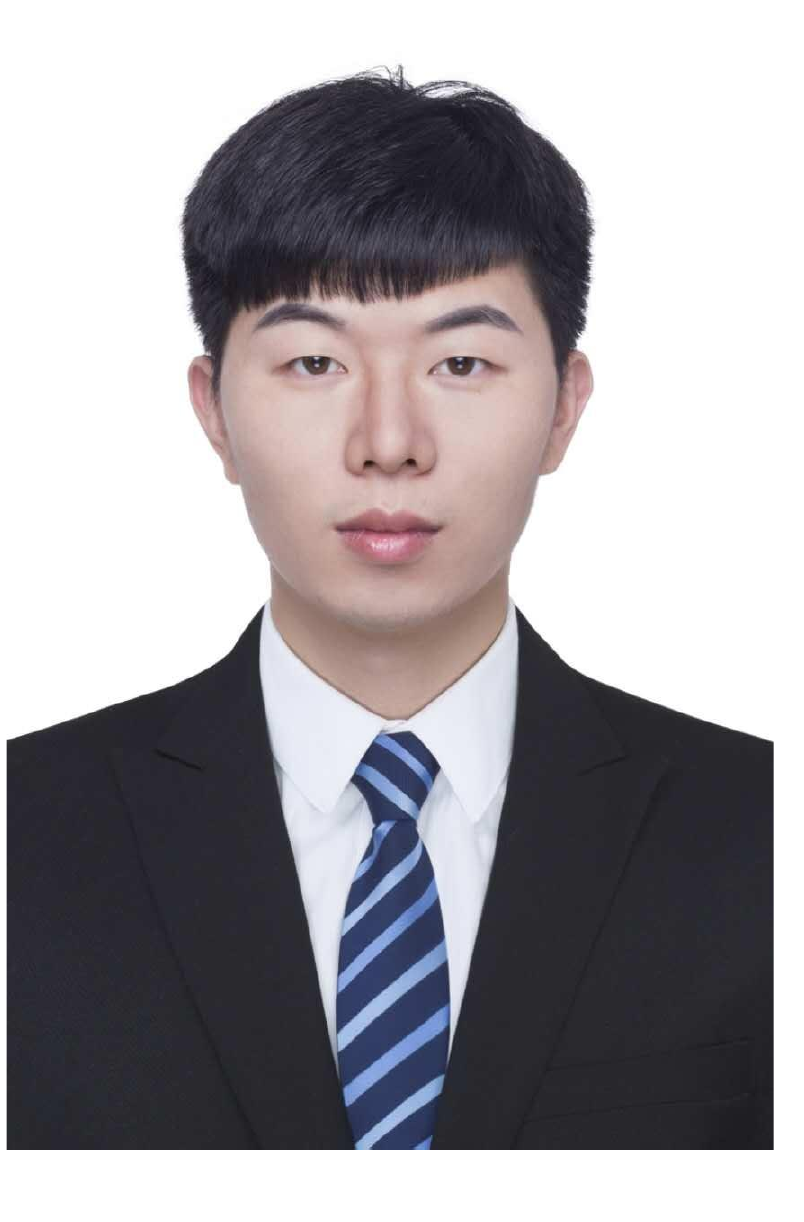}}]{Yan Qin }
	(Member, IEEE) received the B.S. degree in electronic information engineering from Information Engineering University, Zhengzhou, P.R. China, in 2011, the M.S. degree in control theory and control engineering from Northeastern University, Shenyang, P.R. China, in 2013, and the Ph.D. degree in control science and engineering from Zhejiang University, Hangzhou, China, in 2018. From 2019 to 2022, he was a Post-Doctoral Research Fellow with the Singapore University of Technology and Design and the Nanyang Technological University, respectively. He is currently a professor at the School of Automation, Chongqing University, P.R. China. His research interests include data-driven process monitoring, soft sensor modeling, and remaining useful life estimation for industrial processes and essential assets.	
\end{IEEEbiography}
\vspace{-1.5cm}
\begin{IEEEbiography}[{\includegraphics[width=1in,height=1.25in,clip,keepaspectratio]{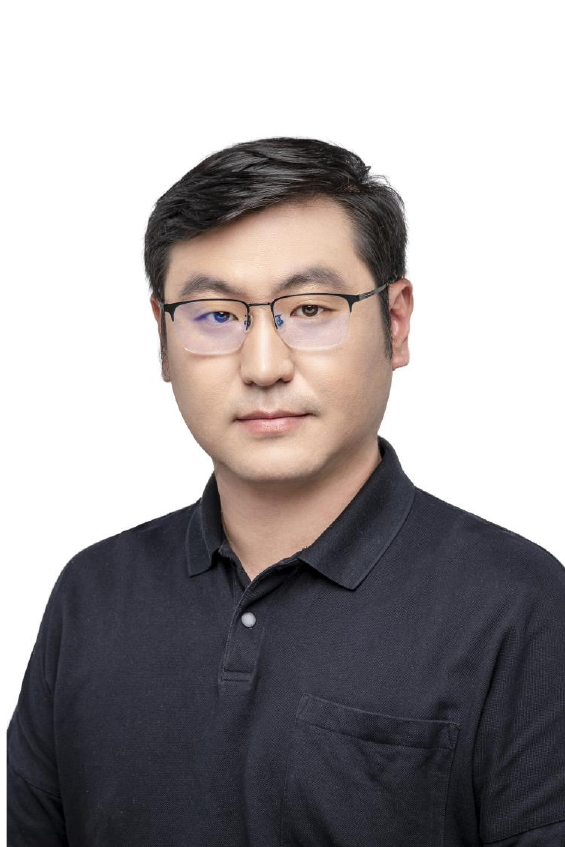}}]{Wei Dai }
	(Senior Member, IEEE) received the Ph.D. degrees in control theory and control engineering from Northeastern University, Shenyang, China, in 2015, respectively. From 2013 to 2015, he was as teaching assistant in the State Key Laboratory of Synthetical Automation for Process Industries, Northeastern University. He is currently a professor and outstanding young backbone teacher in the Engineering Research Center of Intelligent Control for Underground Space, Ministry of Education, China University of Mining and Technology, Xuzhou, China. His current research interests include modeling, optimization and control of complex system, data mining and machine learning, randomized neural network. Prof. Dai won Youth Science and Technology Award of China Coal Society in 2022, the Second Prize of Natural Science Award of the Ministry of Education of China in 2020, the First Prize of Natural Science Award of Chinese Association of Automation in 2021, and the Third Prize of Science and Technology Award of Jiangsu Province in 2019. He also received the Best Paper Award of IEEE International Conference on Real-time Computing and Robotics in 2020. He was a recipient of the Top Young Talents Award of the National Ten Thousand Talents Plan of China in 2021.	
\end{IEEEbiography}
\vspace{-1.5cm}
\begin{IEEEbiography}[{\includegraphics[width=1in,height=1.25in,clip,keepaspectratio]{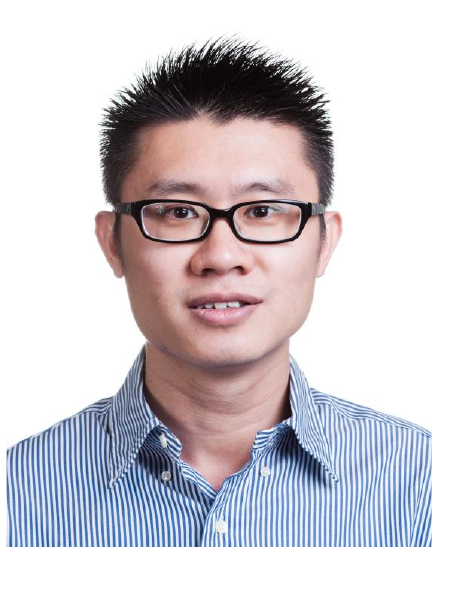}}]{Chau Yuen}
	(S02-M06-SM12-F21) received the B.Eng. and Ph.D. degrees from Nanyang Technological University, Singapore, in 2000 and 2004, respectively. He was a Post-Doctoral Fellow with Lucent Technologies Bell Labs, Murray Hill, in 2005, and a Visiting Assistant Professor with The Hong Kong Polytechnic University in 2008. From 2006 to 2010, he was with the Institute for Infocomm Research, Singapore. From 2010 to 2023, he was with the Engineering Product Development Pillar, Singapore University of Technology and Design. Since 2023, he has been with the School of Electrical and Electronic Engineering, Nanyang Technological University.   Dr. Yuen received IEEE Communications Society Fred W. Ellersick Prize (2023), IEEE Marconi Prize Paper Award in Wireless Communications (2021), and EURASIP Best Paper Award for JOURNAL ON WIRELESS COMMUNICATIONS AND NETWORKING (2021). He was a recipient of the Lee Kuan Yew Gold Medal, the Institution of Electrical Engineers Book Prize, the Institute of Engineering of Singapore Gold Medal, the Merck Sharp and Dohme Gold Medal, and twice a recipient of the Hewlett Packard Prize. He received the IEEE Asia Pacific Outstanding Young Researcher Award in 2012 and IEEE VTS Singapore Chapter Outstanding Service Award on 2019.   Dr Yuen current serves as an Editor-in-Chief for Springer Nature Computer Science, Editor for IEEE TRANSACTIONS ON VEHICULAR TECHNOLOGY, IEEE SYSTEM JOURNAL, and IEEE TRANSACTIONS ON NETWORK SCIENCE AND ENGINEERING, where he was awarded as IEEE TNSE Excellent Editor Award and Top Associate Editor for TVT from 2009 to 2015. He also served as the guest editor for several special issues, including IEEE JOURNAL ON SELECTED AREAS IN COMMUNICATIONS, IEEE WIRELESS COMMUNICATIONS MAGAZINE, IEEE COMMUNICATIONS MAGAZINE, IEEE VEHICULAR TECHNOLOGY MAGAZINE, IEEE TRANSACTIONS ON COGNITIVE COMMUNICATIONS AND NETWORKING, and ELSEVIER APPLIED ENERGY.   He is a Distinguished Lecturer of IEEE Vehicular Technology Society, Top 2\% Scientists by Stanford University, and also a Highly Cited Researcher by Clarivate Web of Science. He has 3 US patents and published over 500 research papers at international journals or conferences.
\end{IEEEbiography}

\end{document}